\documentclass[11pt]{article}
\usepackage{geometry}
\usepackage{coling2020}
\usepackage{times}
\usepackage{url}
\usepackage{latexsym}
\usepackage{microtype}
\usepackage{todonotes}
\usepackage{booktabs}
\usepackage{multirow}
\usepackage{wasysym}
\usepackage{subcaption}

\newcommand{\Sref}[1]{\S\ref{#1}}

\newcommand{\Fref}[1]{Figure~\ref{#1}}

\newcommand{\Tref}[1]{Table~\ref{#1}}

\newcommand{\offenseval}{OffensEval 2020 }

\colingfinalcopy 

\title{NLPDove at SemEval-2020 Task 12: Improving Offensive Language Detection with Cross-lingual Transfer}

\author{Hwijeen Ahn \\
  Sogang University \\
  Seoul, Korea \\
  {\tt hwijeen@sogang.ac.kr} \\\And
  Jimin Sun \\
  Seoul National University \\
  Seoul, Korea \\
  {\tt jiminsun@dm.snu.ac.kr} \\\And
  Chan Young Park \\
  Carnegie Mellon University \\
  Pittsburgh, USA \\
  {\tt chanyoun@cs.cmu.edu}}
  
\author{
  Hwijeen Ahn\textsuperscript{1*}, 
  Jimin Sun\textsuperscript{2*}, 
  Chan Young Park\textsuperscript{3*}, 
  Jungyun Seo \textsuperscript{1}\\
  \textsuperscript{1}Sogang University, Republic of Korea\\
  \textsuperscript{2}Seoul National University, Republic of Korea\\
  \textsuperscript{3}Language Technologies Institute, Carnegie Mellon University, USA\\
  {\tt \{hwijeen, seojy\}@sogang.ac.kr}, \\
  {\tt jiminsun@dm.snu.ac.kr},
  {\tt chanyoun@cs.cmu.edu} \\
}
\date{}

\begin{document}
\maketitle
\begin{abstract}
  This paper describes our approach to the task of identifying offensive languages in a multilingual setting.
  We investigate two data augmentation strategies: using additional semi-supervised labels with different thresholds and cross-lingual transfer with data selection. Leveraging the semi-supervised dataset resulted in performance improvements compared to the baseline trained solely with the manually-annotated dataset. We propose a new metric, Translation Embedding Distance, to measure the transferability of instances for cross-lingual data selection. 
  We also introduce various preprocessing steps tailored for social media text along with methods to fine-tune the pre-trained multilingual BERT (mBERT) for offensive language identification.
  Our multilingual systems achieved competitive results in Greek, Danish, and Turkish at OffensEval 2020.
\end{abstract}

\section{Introduction}
\label{intro}
%
%
\blfootnote{
    %
    %
    %
    %
    %
    \hspace{-0.65cm}  
    This work is licensed under a Creative Commons 
    Attribution 4.0 International License.
    License details:
    \url{http://creativecommons.org/licenses/by/4.0/}.\\
    $\ast$ The first three authors contributed equally.
}

Online social media has become one of the most important means of communication. Unfortunately, the discourse is often laden with abusive language that can have damaging effects on social media users \cite{zampieri2019predicting}.
As online abuse grows as a serious social problem, many social media platforms are employing human moderators to track inappropriate contents and comments.
However, such human moderation comes at the cost of additional operation and the mental suffering of the workers \cite{roberts2014behind,dosono2019moderation}.
In response, the automatic identification of offensive language has been a rising area of interest to both academia and industry.
Computational methods presented in several prior studies  \cite{park-fung-2017-one,DBLP:journals/corr/abs-1801-04433,wulczyn2017ex,founta2018large} have shown promising results on how we can automatically detect a large number of offensive contents with minimal human intervention.

As an effort to promote this line of research, OffensEval was organized at SemEval-2019 \cite{zampieri2019semeval}. 
The organizers released a dataset, Offensive Language Identification dataset (OLID), that contains English tweets with human annotations over three sub-tasks: 1) whether a tweet is offensive or not, 2) whether an offensive tweet is targeted or not, and if so 3) whether the target is an individual or a group.
The second edition of OffensEval, \offenseval: Multilingual Offensive Language Identification in Social Media, holds the same task but extends the OLID to multiple languages (English, Arabic, Danish, Greek, and Turkish) \cite{zampieri-etal-2020-semeval}.

This paper describes our system, NLPDove, submitted to \offenseval. We present two different data augmentation strategies. First, for English, we augment the training set with reliable samples filtered out from the provided semi-supervised dataset. Second, we use samples from languages other than the target language to enhance performance in the target language, namely cross-lingual transfer. Specifically, we propose a new metric, Translation Embedding Distance(\texttt{TED}), to measure the data transferability of each example in the data. 
We use the metric to effectively select additional training data samples from other available transfer languages. 
We also introduce various preprocessing steps to extract meaningful information from user-generated texts despite their inherently noisy nature. Our model is based on the multilingual pre-trained language model, particularly multilingual BERT (mBERT) \cite{devlin-etal-2019-bert}. Our final submitted model is an ensemble of multiple mBERT models with different pooling mechanisms, representation layers and randoms seeds.
Our submission achieved macro F1 of 0.85 in Greek (1st/38), 0.79 in Danish (3rd/40), 0.80 in Turkish (5th/47), 0.91 in English (15th/86), and 0.80 in Arabic (20th/54). 
\footnote{The code for our system can be found at \url{https://github.com/hwijeen/OffensEval2020}}

In the remainder of this paper, we first describe the dataset and the task of \offenseval (\Sref{sec:data}).
We then present our method, including the data augmentation strategies, preprocessing pipeline, and model architecture (\Sref{sec:method}).
The evaluation results of our final submission are reported along with other experiments to further investigate the effectiveness of our method (\Sref{sec:results}).
We then discuss our findings and negative results to shed light on the limitations and remaining challenges in the current offensive language identification models (\Sref{sec:analysis_discussion}).
Finally, we review related work (\Sref{sec:rw}) and conclude (\Sref{sec:conclusion}).

\section{Data}
\label{sec:data}
The most noticeable update in \offenseval is the extension of the task to four additional languages other than English: Arabic \cite{mubarak2020arabic}, Danish \cite{sigurbergsson2020offensive}, Greek \cite{pitenis2020}, and Turkish \cite{coltekikin2020}.
This change facilitates the research of offensive language identification in multiple languages, and also enables researchers to explore the feasibility of cross-lingual approaches of the task.
One thing to note is that unlike English tweets which had annotations for all three sub-tasks, the four new languages do not have all of them; they are only annotated on the first sub-task, which tells whether they are offensive or not.

In the case of English, in addition to the manually annotated tweets from the previous workshop \cite{zampieri2019predicting}, \offenseval released a large semi-supervised dataset where samples are labeled by a set of trained models, not by human annotators \cite{rosenthal2020}. 
Therefore, the labels are not binary but given as floating-point numbers between 0 and 1 (i.e. the average score of multiple models), with their standard deviation indicating the confidence level among classifiers. For further details regarding the data, we refer to the original paper \cite{rosenthal2020}. The summary statistics of the dataset are presented in \Tref{tab:data_stat}.

\section{Method}
\label{sec:method}
There are three important components in our system: training data augmentation, extensive preprocessing, and model ensembling. The following subsections describe each part in detail.

\begin{table}[t]
\begin{minipage}{.5\linewidth}
\centering
\begin{tabular}{@{}lccc@{}}
\toprule
Language    & Train size & Valid size  & OFF:NOT    \\ \midrule
Arabic      &    7202    &   800       &    0.25  \\
Danish      &   2666     &    296      &    0.15 \\
English     &   12691    &     1411    &    0.49  \\
Greek       &   7871     &      874    &    0.40     \\
Turkish     &   28582    &     3176    &    0.24  \\
English-semi & 9M & - & - \\
\bottomrule
\end{tabular}
\caption{Summary statistics of the dataset. English-semi denotes the semi-supervised English dataset. For English, we also used the OLID from OffensEval 2019 \cite{zampieri2019predicting}. We randomly split the provided data into train and validation set by 9:1. }
\label{tab:data_stat}
\end{minipage}\quad
\begin{minipage}{.47\linewidth}
\centering
\begin{tabular}{@{}lcc@{}}
\toprule
Preprocessing               & \# of Cases  \\ \midrule
Emoji substitution          &    1454               \\
Hashtag segmentation        &             2290      \\
Letter casing normalization &             12665      \\
URL replacement             &             2140      \\
Punctuation trimming        &            504       \\ \bottomrule
\end{tabular}
\caption{Preprocessing methods and the number of examples modified by each method in the English training data.}
\label{tab:preprocessing}
\end{minipage}
\end{table}

\subsection{Data Augmentation}
mBERT and other large-scale pre-trained language models are known for their massive number of parameters \cite{conneau2019cross}. 
Thus, they require a substantial amount of data to be fine-tuned sufficiently to the downstream task. 
And yet, the dataset in hand may not be large enough, especially for some languages due to the expensive annotation process. 
To alleviate this data sparsity issue, we propose two data augmentation approaches to make most of the provided semi-supervised labels and multilingual data.

\paragraph{Semi-supervised Labels} 
It has become a common practice to augment training data with machine annotated or synthetic labels when the training data is insufficient \cite{sennrich-etal-2016-improving}. 
There is almost an infinite number of unlabeled Twitter data available, many of which contain offensive languages. 
One way to obtain a substantial amount of pseudo-labels is to predict unlabeled tweets with models trained with existing labeled data. 
\offenseval released over nine million English tweets labeled in this semi-supervised manner \cite{rosenthal2020}.
Each sample was given with its average label probability from several trained models and the standard deviation of the probabilities which indicates the confidence level.
One practical question lies in how to utilize this data while reducing the effect of noisy labels and controlling the data quality.
One straightforward approach is to set certain thresholds to filter out the uncertain examples.
We investigate different thresholding strategies with respect to the two measures, the average probability and the standard deviation.

\paragraph{Cross-lingual Data Selection}
Multilingual pre-trained language models allow seamless cross-lingual transfer; they map texts in different languages into the same representation space without introducing extra costs. 
Cross-lingual transfer can be especially helpful when the training data is limited for the target language, i.e., low-resource settings.
Transferring examples from high-resource languages to low-resource languages often results in great improvements \cite{DBLP:journals/corr/ChenASWC16,DBLP:journals/corr/XuY17a}.

Although it is clear that cross-lingual transfer can be of help, the question of which data to additionally use remains an open problem.
Should we use the entire dataset, or is there a subset better than the entire dataset?
We hypothesize that not all samples in the same transfer language will be equally helpful to the target language task, leading to the conclusion that a carefully selected subset will give higher performance gain than using the whole dataset.

To address the \textit{how} part of data selection, we propose a new metric, translation embedding distance (\texttt{TED}), to quantify the \textit{transferability} of a data sample or a language.
We first define a sample's transferability as how useful the transfer language sample is when transferred to the target language task.
Here, we hypothesize that transferability as a concept correlates with \textit{translatability}; 
if a sentence is easily translated to the target language, it is more likely to be a useful training sample.
To measure \texttt{TED}, we follow the three steps: first, we (machine) translate the training samples in transfer languages into the target language.\footnote{We used Google Translate in our experiments.} 
Then, we extract sentence embeddings for both the original sentence and its translation using mBERT. More specifically, we applied mean pooling over the second-to-last (penultimate) layer of mBERT and used it as the sentence embedding.
Finally, we measure the L2 distance between the two representations and refer to it as the \texttt{TED} of the instance.
The transferability of a \emph{language}, which we define as language-level \texttt{TED}, is simply the average of instance-level \texttt{TED} of samples in the transfer language.

We can then rank transfer samples according to their \texttt{TED} scores and choose top-$k$ samples with the lowest distance. 
Likewise, to find the most transferable languages, we can select the top-$k$ languages based on their language-level \texttt{TED} scores.
To verify our hypothesis that adding samples in other languages can enhance performance in the target language task, we investigate the effectiveness of both language-level and instance-level data selection in \Sref{sec:dataaug}.

\subsection{Preprocessing}
User-generated contents are inherently noisy, whereas the pre-trained language models that we use throughout our system are mostly trained with clean texts such as Wikipedia articles and BookCorpus \cite{devlin-etal-2019-bert}.
To minimize the discrepancy, we applied various preprocessing steps to normalize the noisy tweets provided in our dataset.
In particular, we focus on normalizing the unique conventions of tweets (e.g. hashtags, mentions, emojis) and their irregular usage of punctuation and capitalization.
\Tref{tab:preprocessing} lists all of the considered methodologies and the number of cases found for each method in the English data.
Each item is explained in more detail below.

\paragraph{Emoji Substitution}
Although emojis are effective means of expressing sentiments, learning a good representation of emojis with neural models can be challenging due to their low frequency.
Another challenge is that the most pre-trained language models are not acquainted with the use of emojis as the texts used to train them did not contain emojis.
To fully benefit from the knowledge embedded in pre-trained language models, we replaced all emojis with their plain text descriptions (e.g. \smiley{} $\rightarrow$ smiley face) using an off-the-shelf library\footnote{\url{https://github.com/carpedm20/emoji}}.

\paragraph{Hashtag Segmentation}
Hashtags are prevalent in tweets, and they often contain essential information and keywords.
Conventionally, hashtags are prefixed with the \# symbol followed by phrases concatenated without whitespaces. 
In order to extract all relevant information from hashtags, we split them into words using a word segmentation toolkit\footnote{\url{http://www.grantjenks.com/docs/wordsegment/}}. 
We preserved the \# symbol in the beginning of segmented words to help the model differentiate hashtags from plain texts. 

\paragraph{Letter Casing Normalization}
In social media texts, people often use capitalization to express the intensity of their emotions.
For example, ``GET OUT!'' sounds much more intense than ``get out!''.
Although ``GET'' and ``get'', in this case, possess the same meaning and only differ in the degree of emphasis, they will be assigned to completely different tokens if we do not use a canonical form representing both variants.
Keeping the original form may be useful to preserve the richness in expressing emotional intensity, but the embedding quality of the less frequent variants cannot be assured. In this case, ``GET'' would occur much infrequently than ``get'' in the training corpus.
Therefore, we first normalized these casing variations by lowercasing all words. But at the same time, we prepended a special token \texttt{<has\_cap>} when the word contains an uppercase alphabet and \texttt{<all\_cap>} when the entire word was uppercased.

\paragraph{URL Replacement}
Tweets often include raw urls that redirect users to a reference page. 
In the provided data, most of the urls were already replaced with two special tokens, \texttt{HTTP} and \texttt{URL}. 
We unified them into \texttt{HTTP}, since \texttt{HTTP} is the token used during the pre-training stage of mBERT.

\paragraph{Punctuation Trimming}
One characteristic of social media text is the irregular usage of punctuation marks. For instance, the two sentences ``I am sad.." and ``I am sad........" differ slightly in their use of punctuation marks.
The difference between the two sentences is negligible for a human reader in terms of sentiment. However, their sentence representations may differ significantly since the latter's representation would be diluted with its repeated use of periods. Hence, we limit the maximum number of consecutive punctuation marks to three occurrences.

\subsection{Model}
\label{sec:model}
In our final submission, we used an ensemble of several different mBERT models. 
Here, we explain how we selected and ensembled the models.

\paragraph{mBERT}
mBERT is a multilingual extension of BERT which shares the same architecture with BERT  \cite{devlin-etal-2019-bert}. 
The main difference is that mBERT was trained on the monolingual corpora in multiple languages with an extended vocabulary, which enables seamless cross-lingual transfer. The model can take inputs in various languages, and we can either jointly train a model with both the transfer and target language or perform zero-shot transfer where the model is trained only with the transfer language.
mBERT is shown to be highly effective in both settings \cite{Pires2019HowMI}, implying the model's ability to learn a universal representation space that is shared across languages. 

We took advantage of the multilinguality of mBERT and used it as the main model for our system for all languages.
We added a pooling layer and a linear classifier on top of mBERT for the classification task. 
The pooling layer takes mBERT representations of each word and combines them into a fixed-size vector. \footnote{We experimented with mean pooling, max pooling, concatenation of both, and CNN pooling.}
We used the same hyperparameters values provided in \newcite{Wolf2019HuggingFacesTS} except for the learning rate, random seed, and dropout probability. 
For the three hyperparameters we tuned, we ran a grid-search to select the best configuration based on the validation set performance.
We note that tuning the random seed, as found in \newcite{Dodge2020FineTuningPL}, resulted in a significant performance gain.

\paragraph{Ensemble}
Model ensemble methods are widely employed to improve a model's robustness and generalization performance. 
With this motivation in mind, we developed a pipeline for model ensembling. 
We trained multiple BERT-based models for each task that differ by which BERT layer is used for the word representation (last, second-to-last)\footnote{It has been reported that BERT outputs from different layers encode different kinds of information \cite{kovaleva-etal-2019-revealing}. }, pooling mechanism (mean, max, concatenation of both, and CNN), learning rate (1e-5, 1e-6), and also random seeds. We then chose the top three models according to the validation performance. In the case of English, we additionally considered an mBERT model trained with the augmented dataset that incorporates the semi-supervised dataset as described in \Sref{semi-data}. We used majority voting to aggregate the prediction of the selected models for the final prediction.

\section{Experiment Results}
\label{sec:results}

\begin{table}[]
    \centering
    \begin{tabular}{c|c|c|c|c|c}
        \toprule
        \multirow{2}{*}{Model}&\multicolumn{5}{|c}{Language}\\
         & Arabic & Danish & English & Greek & Turkish \\
        \midrule
        mBERT (Single model) & 0.8722 & \textbf{0.7837} & 0.7770 & \textbf{0.8349} & 0.7835\\
        Ensemble & \textbf{0.8775} & 0.7514 & \textbf{0.7822} & 0.8343 & \textbf{0.7970} \\
        \bottomrule
    \end{tabular}
    \caption{Performance of our system on development data. The ensemble model was used for final submission for all languages.}
    \label{tab:main_result}
\end{table}

\subsection{Main Results}
We used a macro-averaged F1 score as our main evaluation metric as in OffensEval 2020.
Table~\ref{tab:main_result} summarizes the performance of our models in the validation data for each target language.
We also report the best performance of a single mBERT model for comparison.
As explained in \Sref{sec:model}, our final submission models used a majority vote ensemble of several mBERT models. 

\subsection{Data Augmentation Experiment}
\label{sec:dataaug}

\paragraph{Semi-supervised Labels} 
\label{semi-data}
For the English task, we used the human-annotated dataset from OffensEval 2019 as the primary data source and augmented it with the subset of the semi-supervised dataset from OffensEval 2020. 
As explained in Section \ref{sec:data}, the semi-supervised labels are the averaged prediction scores of various trained models accompanied by the standard deviation of these scores. 

One way to control the quality of the additional data is to filter out noisy data using the prediction confidence score of classifiers and their standard deviation.
For instance, setting the minimum value of the offensive class confidence as 0.9 rather than 0.6 would result in a smaller but more reliable subset.
Similarly, a sample with a lower standard deviation indicates a higher likelihood of multiple models agreeing to its predicted label, thereby more reliable.
We performed a grid search over combinations of threshold values for offensive and non-offensive tweets as well as the standard deviation. In particular, we set the grid with three different threshold parameters:  $t_{\mathrm{off}} \in \{0.8, 0.9\}, t_{\mathrm{not}} \in \{0.2, 0.3\}, t_{\mathrm{std}} \in \{0.1, 0.125\}$. 
Data samples are selected when the standard deviation is lower than $t_{\mathrm{std}}$, and its score is either higher than $t_{\mathrm{off}}$ or lower than $t_{\mathrm{not}}$.

The results are summarized in Table~\ref{tab:dist_aug}. 
Although varying the threshold resulted in additional data of different qualities and sizes, the difference in results was marginal. 
When about 30K samples were added with cut-off values of $t_{\mathrm{off}}=0.8, t_{\mathrm{not}}=0.2, t_{\mathrm{std}}=0.125$, we observed a small performance gain compared to others.
In our final ensemble model, we included the models that performed better than the baseline which was trained solely on the human-annotated dataset.

\paragraph{Cross-lingual Data Transfer}
We first consider language-level cross-lingual transfer strategies. That is, we seek the most effective transfer language for the given target language. To this end, we measure language-level \texttt{TED} for each transfer and target language pair. We then pick top-\textit{k} languages that have high language-level \texttt{TED} score for each target language.

Cross-lingual transfer has been shown to be effective especially when the target language dataset is not sufficient. Our experiments focus on the setting where Danish is the target language as it has the least number of training samples. $k$ was set to 1 for language selection. Results showed that English (9.83) is the most useful transfer language to Danish, followed by Arabic (10.85), Greek (22.02) and Turkish (23.08). The high transferability from English to Danish is probably due to the two languages' surface-level resemblance, which comes from the shared orthography. We note that English turned out to be the most effective language to Danish even in terms of zero-shot transfer performance (trained only on the source language and tested on the target language). 

Next, we consider the transfer strategy on a more fine-grained instance-level. We construct a subset of the dataset of the most transferable language based on each sample's transferability.
Specifically, we sort the English samples based on their \texttt{TED} to Danish, and choose the samples with the lowest \texttt{TED}.
We investigate three different ways to build the subset: taking top/bottom/random-\textit{k} English samples ($\mathrm{en}_{\mathrm{top}}$, $\mathrm{en}_{\mathrm{bottom}}$, $\mathrm{en}_{\mathrm{rand}}$). 
In our experiment, we set \textit{k} to 1300, which is about half the size of the Danish dataset.
We compare the results with two baselines, one trained only with Danish data ($\mathrm{da}$) and another one trained with the aggregated dataset of Danish and \emph{all} English samples ($\mathrm{da}+\mathrm{en}_{\mathrm{all}}$). The results are in Table \ref{tab:cross_aug}. 

Results confirmed our hypothesis that we can further improve cross-lingual transfer by selecting samples that are more transferable than others.
 $\mathrm{da}+\mathrm{en}_{\mathrm{top}}$ showed the best performance, while $\mathrm{da}+\mathrm{en}_{\mathrm{bottom}}$ and $\mathrm{da}+\mathrm{en}_{\mathrm{rand}}$ performed worse than $\mathrm{da}+\mathrm{en}_{\mathrm{all}}$.
This shows that na\"ively using all samples in the transfer language (English) is not necessarily the best way to conduct cross-lingual transfer.

\begin{table}[]
    \centering
            \begin{tabular}{c|c|c|c|c}
        \toprule
        \multicolumn{3}{c|} {Threshold} & \multirow{1}{*}{Additional} & \multirow{1}{*}{Macro} \\
         \multicolumn{1}{c}{OFF} & \multicolumn{1}{c}{NOT} & \multicolumn{1}{c|}{std} & data size & F1 \\
        \midrule
        0.8 & 0.2 & 0.1 & 3533 & 0.7693\\
        0.8 & 0.2 & 0.125 & 29681 & \textbf{0.7793}\\
        0.8 & 0.3 & 0.1 & 7956 & 0.7716\\
        0.8 & 0.3 & 0.125 & 83782 & 0.7738\\
        0.9 & 0.2 & 0.1 & 2769 & 0.7741\\
        0.9 & 0.2 & 0.125 & 19590 & 0.7709 \\
        0.9 & 0.3 & 0.1 & 7192 & 0.7707 \\
        0.9 & 0.3 & 0.125 & 73691 & 0.7453\\
        \midrule
        \multicolumn{3}{c|} {Baseline} & 0 & 0.7754\\
        \bottomrule
    \end{tabular}
    \caption{Results of data augmentation from semi-supervised labels. F1 scores are measured over the validation split of English data.}
    \label{tab:dist_aug}
\end{table}

\begin{table}[]
    \centering
    \begin{tabular}{c|c|c|c|c|c}
        \toprule
         Train data & $da$ & $da+en_{all}$ & $da+en_{top}$ & $da+en_{bottom}$ & $da+en_{rand}$ \\
        \midrule
        Macro F1 & 0.77 & 0.78 & \textbf{0.84} & 0.72 & 0.72 \\ \bottomrule
    \end{tabular}
    \caption{Results of cross-lingual data augmentation on the Danish validation data. $da$ refers to the original Danish training data, and $\mathrm{en}_{\mathrm{all}}$ is the entire English training dataset. $\mathrm{en}_{\mathrm{top}}$, $\mathrm{en}_{\mathrm{bottom}}$, $\mathrm{en}_{\mathrm{rand}}$ are the subsets of samples according to the \texttt{TED}.}
    \label{tab:cross_aug}
\end{table}

\section{Analysis and Discussion}
\label{sec:analysis_discussion}
In this section, we analyze how cross-lingual data selection could help improve performance. In addition, we discuss limitations of the proposed metric, \texttt{TED}, and some negative results with masking offensive words with a hope to provide insights for future work.
\label{sec:discussion}

\subsection{How Does Cross-lingual Data Selection Help?}
We showed that our data augmentation strategy using data from other languages helps improve the target task performance in Section \ref{sec:dataaug}. 
Here, we investigate why cross-lingual data selection was helpful.

\begin{figure*}
    \centering
    \includegraphics[width=0.45\textwidth]{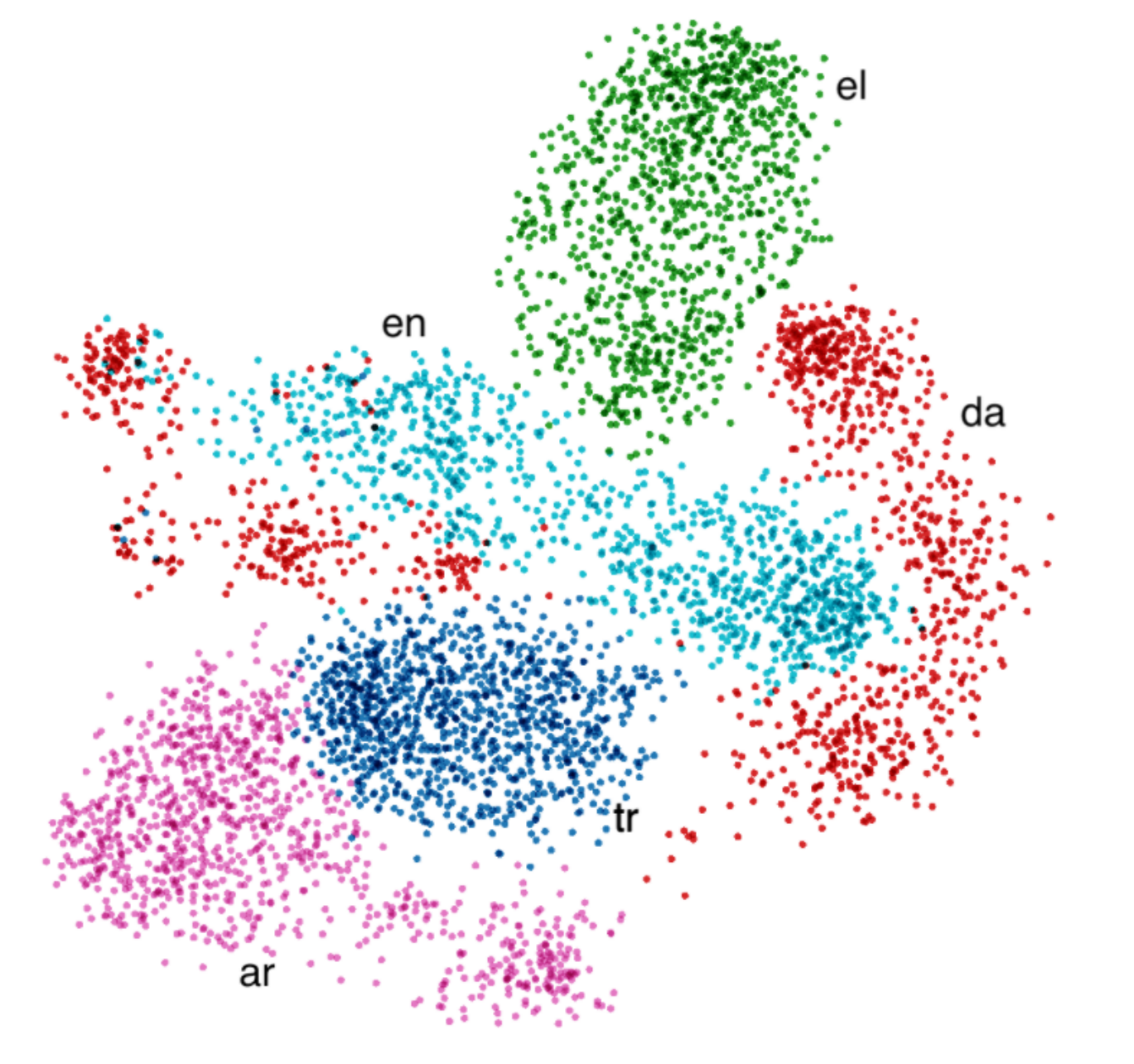}
    \caption{tSNE visualization of mBERT sentence embeddings of OLID samples. Samples are color coded according to their language (ar:Arabic, da:Danish, en:English, el:Greek, and tr:Turkish).}
    \label{fig:tsne_all}
\end{figure*}

We first take a look at the embedding space of mBERT to understand why language-level data selection, according to language-level \texttt{TED}, was beneficial. We randomly sampled a thousand samples from each language and visualized their sentence embeddings using tSNE \cite{maaten2008visualizing} in \Fref{fig:tsne_all}. 
From the visualization, we observed that the examples formed rough clusters based on their language. Interestingly, English and Danish clusters are placed near one another. This aligns with our experimental result in the previous section, where we found English to be the most transferable language to Danish by language-level \texttt{TED}. We speculate that this proximity in the embedding space may be related to the improvement yielded when English samples with low \texttt{TED} were added to the Danish dataset. mBERT might have benefited from the additional English data even while training a Danish classifier because the examples were relatively close to each other in the shared embedding space.

In addition, we manually inspect the examples with low \texttt{TED} scores which we assumed to be more transferable. Examples in \Tref{tab:transf-sample} reveal that the samples with low \texttt{TED} are the ones that did not change significantly after the translation procedure. These samples often contain words used in both languages colloquially (e.g., creepy, fool), and tend to be short with a rather simple grammatical structure.
We posit that this simplistic nature of the samples with low \texttt{TED} made themselves more translatable, thus more transferable, and led to improvements via cross-lingual transfer.

\subsection{Limitations of \texttt{TED}}
\texttt{TED} measures the transferability of an example with the help of a machine translation system. Incorporating an external off-the-shelf translator is convenient in that we do not need to build an additional translation system for multiple pairs of languages. However, still, the computation of \texttt{TED} comes with a cost of translating all samples for all possible transfer-target language pairs. 
Another limitation of our method is that the quality of data selection relies on the performance of the employed machine translation system.
This dependency may lead to the misrepresentation of transferability in low-resource language samples because the quality of translation systems in these languages are generally deficient.
We believe that the different approaches commonly used for cross-lingual transfer (e.g., alignment methods) may pave the way for building a transferability metric that does not rely on external translation modules. 

\subsection{Offensive Words in Inoffensive Contexts}
In our pilot experiment, we fine-tuned BERT with the OLID from OffensEval 2019 \cite{zampieri2019semeval}. The model's initial performance was around 0.80 macro F1, which is comparable considering that the best performance from OffensEval 2019 was 0.82 \cite{liu-etal-2019-nuli}. Qualitative analysis on the false-positive cases, however, showed that the model tends to classify tweets that include offensive words as offensive, regardless of the context. For example, \textit{``@USER And the fact he called you a b*tch after you respectfully told him your preference.."} was classified as offensive although the term ``b*tch" was not used to express any aggression toward the listener. Offensive words are also often used for alternative motives: as a topic of the statement or even to emphasize a positive sentiment.
We have found many cases in the dataset where offensive terms have not been used in a despicable manner (e.g., \textit{``f*cking awesome"}). However, since most offensive words are paired with aggressive intentions, the model's prediction seemed to be dictated by the presence of these words, ignoring the subtle context.

Our attempt to help the model focus more on the context and differentiate these subtle cases was to randomly mask some of the offensive words during training. The intuition was to encourage the model to attend to the context instead of specific terms by partially removing the dependency on certain words. 
During model training, when an input tweet contained words in the offensive word list\footnote{https://www.cs.cmu.edu/~biglou/resources/, https://www.freewebheaders.com/full-list-of-bad-words-banned-by-google, https://github.com/LDNOOBW/List-of-Dirty-Naughty-Obscene-and-Otherwise-Bad-Words}, we randomly replaced them with a \texttt{[PAD]} token. We used the unmasked, original sample at inference time. We expected this procedure to help the model become more robust against different contexts of offensive word usage. Unfortunately, the masking strategy did not result in quantitative gains. A possible explanation for the performance decline could be that the na\"ive masking excessively deleted offensive words that were actually used offensively. A more balanced approach would be required to address this limitation in future work.



\begin{table}[]
\centering
\begin{tabular}{p{20em}|p{20em}}
\toprule
     Original (English) & Translated (Danish) \\
     \midrule
    @USER Creepy & @USER Creepy \\
    \midrule
    @USER Fool! & @USER Fool! \\
    \midrule
    \#RestoreHumanity \#AntiFa September 22: Stop the fascist NVU in \#Amsterdam: URL HT @USER & \#RestoreHumanity \#AntiFa September 22: Stop det fascistiske NVU in \#Amsterdam: URL HT @USER\\
    \midrule
    @USER I'M SO FU*KING READY &  @USER I'M SO FU*KING KLAR\\
    \midrule
    @USER @USER so sad \#taketworeferenceswithmoniqueandchloe x & @USER @USER så trist \#taketworeferenceswithmoniqueandchloe x \\
\bottomrule
\end{tabular}
\caption{English samples with low \texttt{TED} when transferred to Danish. Most examples were not changed significantly after the translation.}
\label{tab:transf-sample}
\end{table}

\section{Related Work}
\label{sec:rw}
Computational methods to identify offensive language have been studied in many different aspects \cite{davidson2017automated,schmidt2017survey}. Researchers have focused on a variety of categories of offensive language such as sexism, racism, and hate speech in general. \newcite{sap-etal-2019-risk} identified a racial bias in hate speech detection datasets and proposed better annotation methods to address the problem. \newcite{park-fung-2017-one} pointed out the gender bias in abusive language detection models and suggested a de-biasing algorithm. \newcite{qian-etal-2018-hierarchical} introduced a more fine-grained hate speech identification task than binary classification, and presented an architecture that leverages Conditional Variational Autoencoder (CVAE).

Large-scale pre-trained models like BERT have achieved state-of-the-art results in various NLP tasks \cite{devlin-etal-2019-bert,yang2019xlnet}. \newcite{wu-dredze-2019-beto}, \newcite{K2020Cross-Lingual}, and \newcite{Pires2019HowMI} explored the multilinguality of mBERT and confirmed the model's effectiveness in cross-lingual transfer.
In the task of offensive language detection,  \cite{liu-etal-2019-nuli} has adopted BERT to solve the problem.
In our work, we also use mBERT but improve the method's ability under a multilingual setting by leveraging cross-lingual transfer.

\section{Conclusion}
\label{sec:conclusion}
In this paper, we describe NLPDove's approach to SemEval-2020 Task 12: Multilingual Offensive Language Identification in Social Media. The system yielded significant results in the competition for multiple languages, confirming the strength of large-scale pre-trained language models and the importance of various preprocessing schemes and model ensembling. We examine two data augmentation strategies and propose a new metric, \texttt{TED}, to quantify cross-lingual transferability. In qualitative analysis, we inspect factors why data selection with \texttt{TED} increased performance. 
We also present the limitations of our system, mainly failing to detect the speaker's intention expressed subtly in the context, hoping to offer valuable insights and elicit future work in the field. 

\bibliographystyle{coling}
\bibliography{coling2020}

\end{document}